# Model Validation of a Low-Speed and Reverse Driving Articulated Vehicle

***V.V. Gosar[1], M. Alirezaei[1,2], I.J.M. Besselink[1], H. Nijmeijer[1]**
**[1]Eindhoven University of Technology, Eindhoven, The Netherlands**
**[2]Siemens Digital Industries Software, Helmond, The Netherlands**
E-mail: v.v.gosar@tue.nl

For the autonomous operation of articulated vehicles at distribution centers, accurate positioning of the vehicle is of the utmost importance. Automation of these vehicle poses several challenges, e.g. large swept path, asymmetric steering response, large slide slip angles of non-steered trailer axles and trailer instability while reversing. Therefore, a validated vehicle model is required that accurately and efficiently predicts the states of the vehicle. Unlike forward driving, open-loop validation methods can't be used for reverse driving of articulated vehicles due to their unstable dynamics. This paper proposes an approach to stabilize the unstable pole of the system and compares three vehicle models (kinematic, non-linear single track and multibody dynamics model) against real-world test data obtained from low-speed experiments at a distribution center. It is concluded that single track non-linear model has a better performance in comparison to other models for large articulation angles and reverse driving maneuvers.

## 1. INTRODUCTION

With the recent changes in customers demand, the logistic organization face an urge to rethink their business strategy and environmental impact. By implementing automated driving of articulated commercial vehicles at a distribution center (DC), the handling of goods can be done more efficiently. To achieve this, research on the automation of these vehicles has received attention from both industrial and academic parties [1].

The first step towards automation is developing a simulation model that predicts the behavior of the vehicle. In a DC, the minimum distance that occurs between vehicle combinations is small and the requirement for the semi-trailer to be positioned accurately with respect to the docking gate is high. Hence, a vehicle model is required that accurately and efficiently predicts the state of the vehicle while driving forward and reversing.

Most recent studies have evaluated the accuracy of vehicle models for forward driving scenarios from an efficiency point of view. Brock et al. [2] compare trailer axle position estimation of reduced-order single-track linear models from four studies in the literature, and a kinematic model against their high-fidelity multi-body model. Brock et al. have shown that Luijten et al's model [3] produces the smallest position error, whereas the kinematic model generates the maximum position error due to the absence of tire slip. Moreover, Ghandriz et al. [4] compare the dynamic behavior of single and two track nonlinear multi-trailer models against experiments. The study concludes that the single track non-linear model shows the smallest RMSE and similar dynamic behavior to the experiment. However, previous studies do not aim at validating the vehicle model for the very low speed (less than 10 km/hr) and large steering angle (greater than 90°) that typically occurs in confined spaces such as DCs.

The main challenge in the automation of articulated vehicles is, reversing during parking/docking maneuvers due to the unstable dynamics of the vehicle. This instability poses a problem to conduct model validation procedures in an open-loop configuration as is usual for forward driving scenarios. Smith and Chen [5] address the problem of model validation for an unstable system using an inverted pendulum setup. Here, the unstable system is first stabilized using a controller and then validated to see whether it satisfies input-output bounds for an unknown, but bounded, input disturbance. In contrary, system identification techniques have been used to validate open-loop unstable systems (e.g. aircraft) via closed-loop control, see Jategaonkar [6]. In the context of articulated vehicles, the open-loop unstable motion has to be stabilized by the driver. To make a fair comparison between measurements and models, we use feedback control to stabilize the articulation angle in this paper.

In summary, the contributions of this paper are twofold. First, the accuracy of vehicle models is assessed for low-speed forward driving and large articulation angle maneuvers. Second, a stabilizing controller has been developed to allow vehicle model validation for reverse driving maneuvers. This paper is organized as follows. Section 2 gives an overview of the experimental setup to obtain measurement data. Section 3 explains the three vehicle models used in this study. Additionally, a stabilizing controller for reverse driving maneuver is presented. This section ends by describing the distance based error criteria used to evaluate the performance of the models. In section 4, results for the forward and reverse driving maneuvers are discussed. The final section gives concluding remarks and points out potential topics for future work.

## 2. EXPERIMENTS

The measurement data has been collected by performing driving maneuvers at the Dynamic Parcel Distribution (DPD) center in Oirschot, the Netherlands, which is representative of a distribution center.

### 2.1 Test Vehicle

The test vehicle, as shown in Fig. 1, is a two-axle tractor and a tautliner semi-trailer with three non-steered axles. The vehicle combination was tested for two different loading conditions. The first condition was a tractor along with an empty trailer. For the second condition, the trailer was loaded with 10 tons of empty pallets equally distributed along the trailer. Measured axle loads for both conditions can be found in the Appendix

### 2.2 Sensor Instrumentation

For localization, both the tractor and semi-trailer are equipped with an OXTS RT3000 inertial navigation system. The system has an integrated IMU to capture the motions of the vehicle. Additionally, it is enabled with RTK GNSS following NTRIP protocol to receive real-time correction to improve position accuracy. The sensors are positioned in the tractor cabin and behind the last axle of the semi-trailer, as shown in Fig. 1. Furthermore, the Tractor CAN bus is used to record actuation signals like the steering wheel angle. The measured data from OXTS and steering actuator have been recorded at a sampling frequency of 100 Hz, whereas the other signal from the CAN bus have been recorded at 10 Hz. Since the focus is on low-speed driving and planar motion of the vehicle, only the following signals will be considered:

- Steering wheel angle from tractor CAN-bus.
- The longitudinal velocity of tractor and trailer from OXTS.
- The lateral velocity of tractor and trailer from OXTS.
- Yaw rate of the tractor and trailer from OXTS.
- Global position and orientation of tractor and trailer from OXTS.

Table 1 lists the position of the OXTS sensor with respect to the drive axle of the tractor, following the ISO sign convention.

### 2.3 Test Procedure

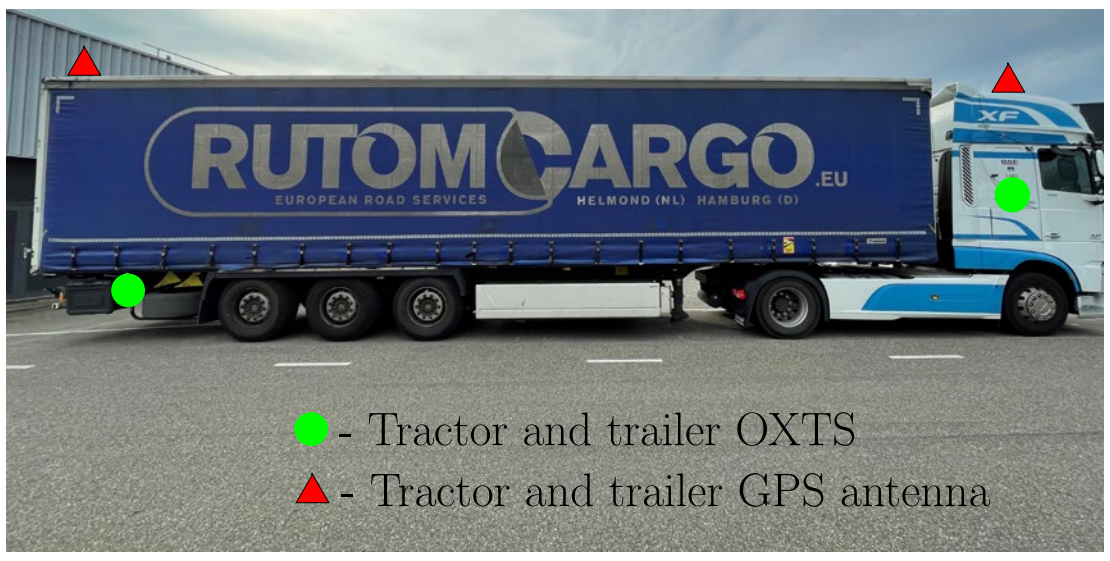

*Fig. 1: Test Vehicle*

*Table 1: Sensor Positions*

| Distance from drive axle [m] | Tractor OXTS | Trailer OXTS |
|---|---|---|
| X | 3.825 | - 10.492 |
| Y | - 0.005 | - 0.004 |
| Z | 1.206 | 0.1 |

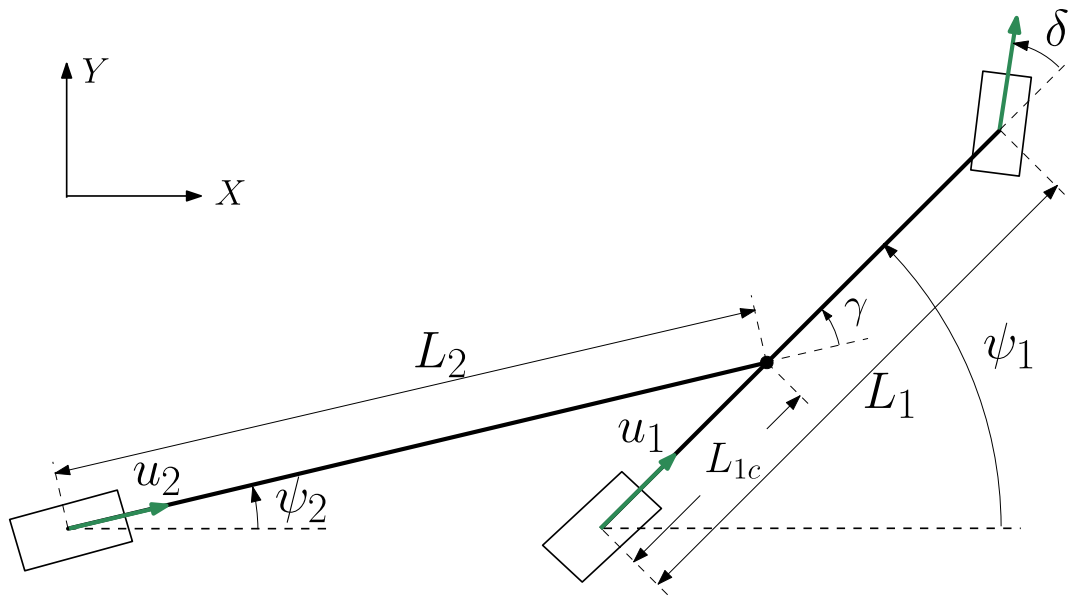

*Fig. 2: Kinematic Model Layout*

A total of 25 forward and 33 reverse driving maneuvers were executed at the DPD distribution center. Forward driving maneuvers included:

- Constant steer tests for 200°, 360° and 500° steering amplitude at 5 and 10 km/hr, including left and right-hand turns.
- For the slalom and figure of 8 test, the vehicle was driven at 5 km/hr.

Reverse driving maneuvers included:

- Ramp steer tests for 30° and 90° steering amplitude at 3 and 6 km/hr, including left and right-hand turn variation. The test was executed before the vehicle reached jackknife condition. The average maximum articulation angle achieved was 45°.
- Docking and parking tests were performed by DC driver with 27 years of experience. On average, maximum articulation angle achieved was 40°.

## 3. VEHICLE MODELLING

In this section, the three used vehicle models (kinematic, non-linear single track and multibody) are described. We aim to use either the kinematic or single track non-linear model for planning and control purposes. The multibody model will be used as the plant in the simulation environment.

### 3.1 Kinematic Model

First, a non-linear single axle kinematic model (KIN) as shown in Fig. 2 is used. The governing equations in the global coordinate frame read:

$$\begin{bmatrix} \dot{X}_1 \\ \dot{Y}_1 \\ \dot{\psi}_1 \\ \dot{\gamma} \end{bmatrix} = \begin{bmatrix} u_1 \cos\psi_1 \\ u_1 \sin\psi_1 \\ u_1 \tan\delta / L_1 \\ -\frac{u_1 \sin\gamma}{L_2} + \left(1 - \frac{L_{1c}\cos\gamma}{L_2}\right)\frac{u_1 \tan\delta}{L_1} \end{bmatrix}, \qquad (1)$$

where $X_1$ and $Y_1$ describe the global position of the tractor drive axle, $\psi_1$ is the tractor heading angle and $\gamma$ equals the articulation angle. Vehicle parameters are in the Appendix.

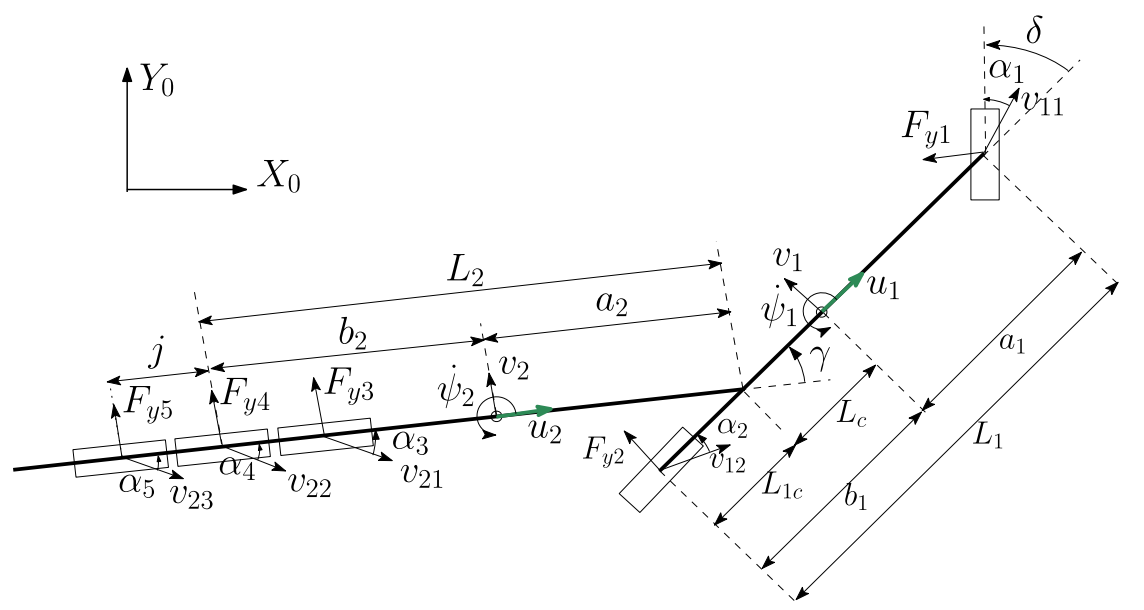

*Fig. 3: STM Layout*

### 3.2 Non-Linear Single Track Model

The second model is a single track non-linear dynamic model (STM) as shown in Fig. 3. The equation of motion is derived using the Lagrangian approach. Local coordinates in the form $\underline{x} = [u_1, v_1, \dot{\psi}_1, \dot{\gamma}]^T$ are used to simplify the expressions for the tire model:

$$M(\gamma)\underline{\dot{x}} + H(\gamma, \underline{x}) = \underline{Q_v} \qquad (2)$$

The linear tire characteristics are modeled by using normalized cornering stiffnesses $f$, Luijten [3]. This implies that axle cornering stiffness scales with the vertical axle load. The matrices $M$, $H$, and $Q_v$ are defined in the Appendix along with the vehicle and tire model parameters. The inputs to the KIN and STM model are wheel angle $\delta$ and longitudinal velocity $u_1$.

### 3.3 Multibody Model

The final model is a 44 degree of freedom (DOF) multi-body tractor semitrailer model (MBD), developed using MATLAB's Simscape Multibody toolbox. The model incorporates tire non-linear behavior using the TNO-Delft tire module, see Evers [7]. In the past, this model has been validated for a wide range of test measurements involving high-speed maneuvers, see Kural [8].

Note, that the outputs from all three models are translated to the OXTS sensor position to allow a direct comparison.

### 3.4 Steering Geometry Modification

A tractor steering system applies steering command to the left front wheel, which is then transferred to the right front wheel via tie-rod, see Loof [9]. This geometry leads to an asymmetric vehicle response while turning the steering wheel to the right and left, specifically when applying large steering angles. This asymmetry is also observed in the vehicle measurements. To accommodate such behavior, the steering input is modeled as a quadratic function:

$$\delta_{center} = \frac{1}{i_s}(\delta_h - q\delta_h^2) \qquad (3)$$

where $\delta_h$ is the steering wheel angle and $i_s$ is the steering ratio and $\delta_{center}$ is the front wheel angle at the center of the front axle. Additionally, when analyzing the relation between steering wheel angle and curvature, about 5° offset of the steering wheel angle $\delta_h$ for zero path curvature was observed. This correction is accounted for in all models.

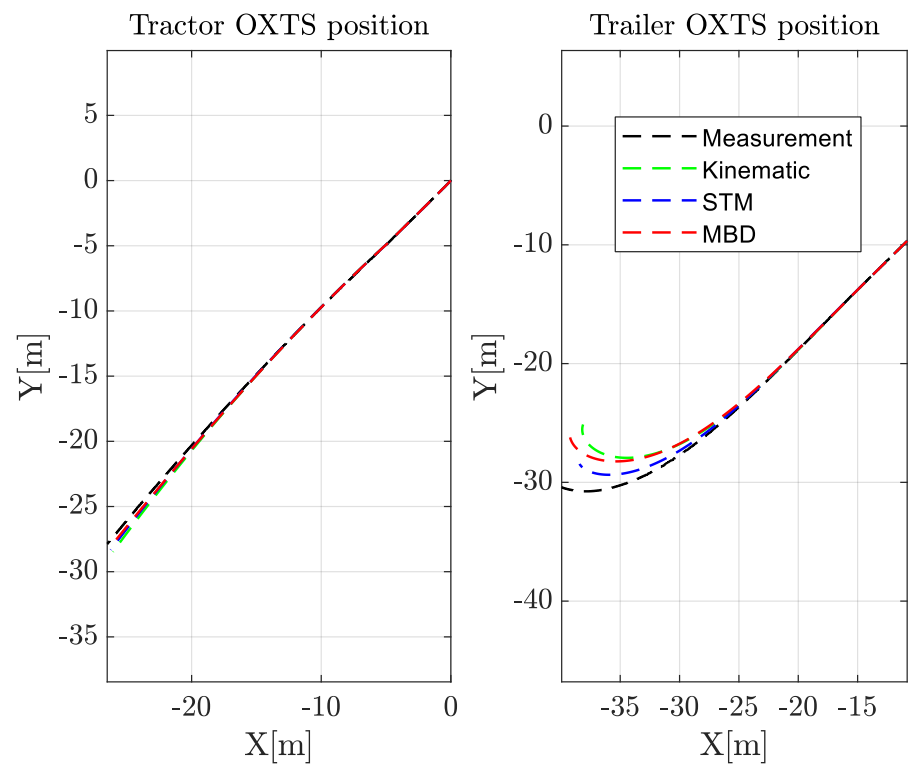

*Fig. 4: Open Loop, Reverse Driving*

### 3.4 Parameter tuning

The vehicle dimensions and weight parameter used in the model are the same as the measured parameter during the test. The parameters tuned are the normalized tire cornering stiffness $f$ for STM model, the steering ratio $i_s$ and steering parameter $q$ for all three models. Using the error criteria defined in section 3.6, the tuning process has been carried out until the models show the smallest error in comparison to the measurement results of the 8 forward driving maneuvers. The parameters of both the unloaded and loaded vehicle can be found in the Appendix. It should be noted that only limited tuning of the MBD model parameters was carried out.

### 3.5 Reverse Driving

Reverse driving of an articulated vehicles results in an unstable system, this can be well explained using the linearized expression for the articulation angle dynamics:

$$\dot{\gamma} = u_1\left(-\frac{\gamma}{L_2} + \left(1 - \frac{L_{1c}}{L_2}\right)\frac{\delta}{L_1}\right) \qquad (4)$$

It can be seen that driving the vehicle backward ($u_1$ <0) with small values of $\delta$ and $\gamma$, a positive eigenvalue is obtained (- $u_1$ /$L_2$) which causes exponential growth of $\gamma$. This can be seen from the semi-trailer position of the models compared to the measurement in Fig. 4 for an open loop simulation. To make a fair comparison between measurements and models, stabilizing controller with proportional feedback action is introduced:

$$\delta = \delta_{center} + K(\gamma_{measured} - \gamma_{model}), \qquad (5)$$

where $\delta_{center}$ is defined in (3), $\gamma_{measured}$ is the measured articulation angle, $\gamma_{model}$ is the articulation angle calculated by the model and $K$ is the proportional feedback gain. Furthermore, using (4) and (5), a lower bound for $K$ can be obtained to ensure stability:

$$\dot{\gamma} = u_1\left(\frac{-L_1 + K(L_2 - L_{1c})}{L_1 L_2}\right)\gamma \qquad (6)$$

For $u_1 < 0$ the following inequality should hold to ensure stability:

$$-L_1 + K(L_2 - L_{1c}) > 0 \qquad (7)$$

Therefore, the minimum gain for the controller to stabilize the vehicle in reverse driving should be $K > L_1/(L_2 - L_{1c})$. In this study, $K = 3$ is used for all three models.

### 3.6 Error Criteria

To compare the three vehicle models with the measurements, the following error for both tractor and trailer unit are defined: lateral position$(\epsilon_{y1}, \epsilon_{y2})$, yawrate $(\epsilon_{\dot{\psi}_1}, \epsilon_{\dot{\psi}_2})$ and lateral velocity $(\epsilon_{vy1}, \epsilon_{vy2})$. The lateral position error is obtained using the vector calculus method to find the orthogonal distance (the shortest distance) between the path generated by the model and measured path. The yawrate and lateral velocity are compared with measured data as a function of time. Thereafter, the obtained error is averaged over the traveled distance to avoid dependency on speed and to make sure longer measurements do not generate a large error. The following equation is used.

$$\bar{\epsilon} = \sqrt{\frac{1}{s_{max}} \int_0^{s_{max}} \epsilon^2 \, ds} \qquad (8)$$

The absolute error for the vehicle for lateral position, yaw rate and lateral velocity is defined by:

$$\begin{aligned} \bar{\epsilon}_p &= \bar{\epsilon}_{y1} + \bar{\epsilon}_{y2} \ , \\ \bar{\epsilon}_a &= \bar{\epsilon}_{\dot{\psi}1} + \bar{\epsilon}_{\dot{\psi}2} \ , \\ \bar{\epsilon}_v &= \bar{\epsilon}_{vy1} + \bar{\epsilon}_{vy2} \end{aligned} \qquad (9)$$

Further, the absolute error $\bar{\epsilon}_a$ and $\bar{\epsilon}_v$ are normalized by using mean of their respective measured signal. Finally, their summation is the total normalized error $\bar{\epsilon}_n$. For reverse driving the models are also evaluated based on the normalized steering effort required to stabilize the vehicle:

$$J_{steer} = \frac{\sqrt{\frac{1}{s_{max}} \int_0^{s_{max}} |\delta - \delta_{measured}|^2 \, ds}}{|\delta_{measured}|} \qquad (10)$$

where $\delta$ is defined in (5) and $\delta_{measured}$ is the measured steering wheel angle.

## 4. RESULTS AND DISCUSSION

In this section, the results of the three models developed are first compared against measurement data for forward driving maneuvers and reverse driving maneuvers. Thereafter, visual representations of the states and vehicle position are discussed for both cases. The overall accuracy of the models is tabulated based on the error criterion defined in section 3.6.

### 4.1 Forward Driving Results

For a qualitative analysis, results of the constant steer test performed at a steering amplitude of 200° and velocity of 5km/h are shown in Fig. 5. Here, the first two subplots show the steering wheel angle and velocity input to the models. The next four subplots show the lateral velocity and yawrate of the tractor and trailer respectively. It can be observed that the STM and MBD models accurately estimate the lateral velocity and yaw rate response. Note, the steering ratio is tuned to find the best fit for the model, hence they are different for all the models.

For quantitative analysis, the results of the 17 forward driving maneuvers are listed in Table 2. It lists the errors defined section 3.6. For the forward driving scenario, model assessment factors are $\bar{\epsilon}_n$ and $\bar{\epsilon}_p$. It can be concluded the MBD model produces smallest error (4.83 %). Since these are open loop tests without any correction for drifts, the overall position accuracy by all the models is still low.

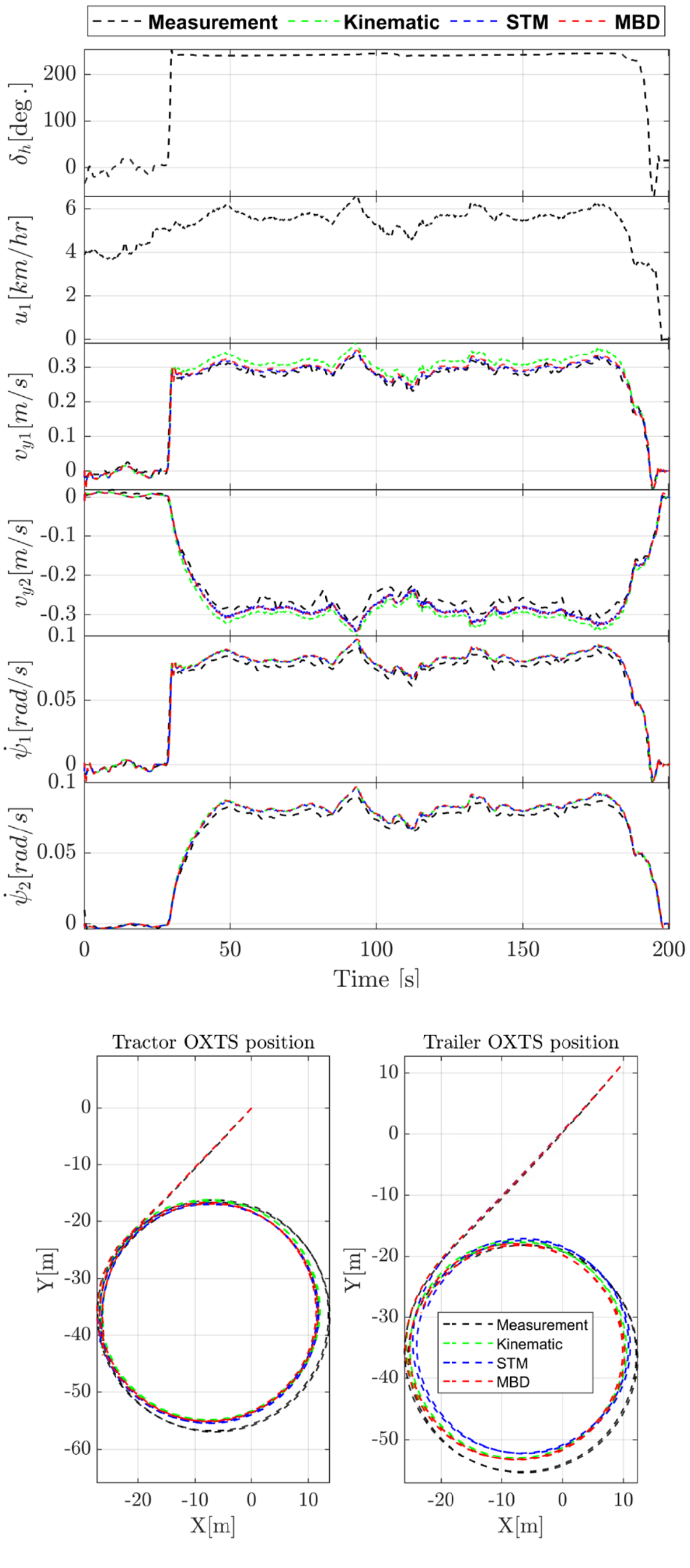

*Fig. 5: Constant Steer Test*

*Table 2: Forward Driving Results*

| Error | $\bar{\epsilon}_\alpha$ rad/s | $\bar{\epsilon}_v$ m/s | $\bar{\epsilon}_p$ m | $\bar{\epsilon}_n$% |
|---|---|---|---|---|
| *KIN* | 0.0064 | 0.0431 | 3.06 | 9.26 |
| *STM* | 0.0049 | 0.0211 | 2.51 | 5.9 |
| *MBD* | 0.0037 | 0.0159 | 1.72 | 4.83 |

*Table 3: Reverse Driving Results*

| Error | $\bar{\epsilon}_\alpha$ rad/s | $\bar{\epsilon}_v$ m/s | $J_{steer}$[%] | $\bar{\epsilon}_n$ [%] |
|---|---|---|---|---|
| *KIN* | 0.0072 | 0.039 | 26.96 | 32.36 |
| *STM* | 0.0069 | 0.032 | 21.15 | 27.7 |
| *MBD* | 0.0080 | 0.034 | 26.14 | 30.78 |

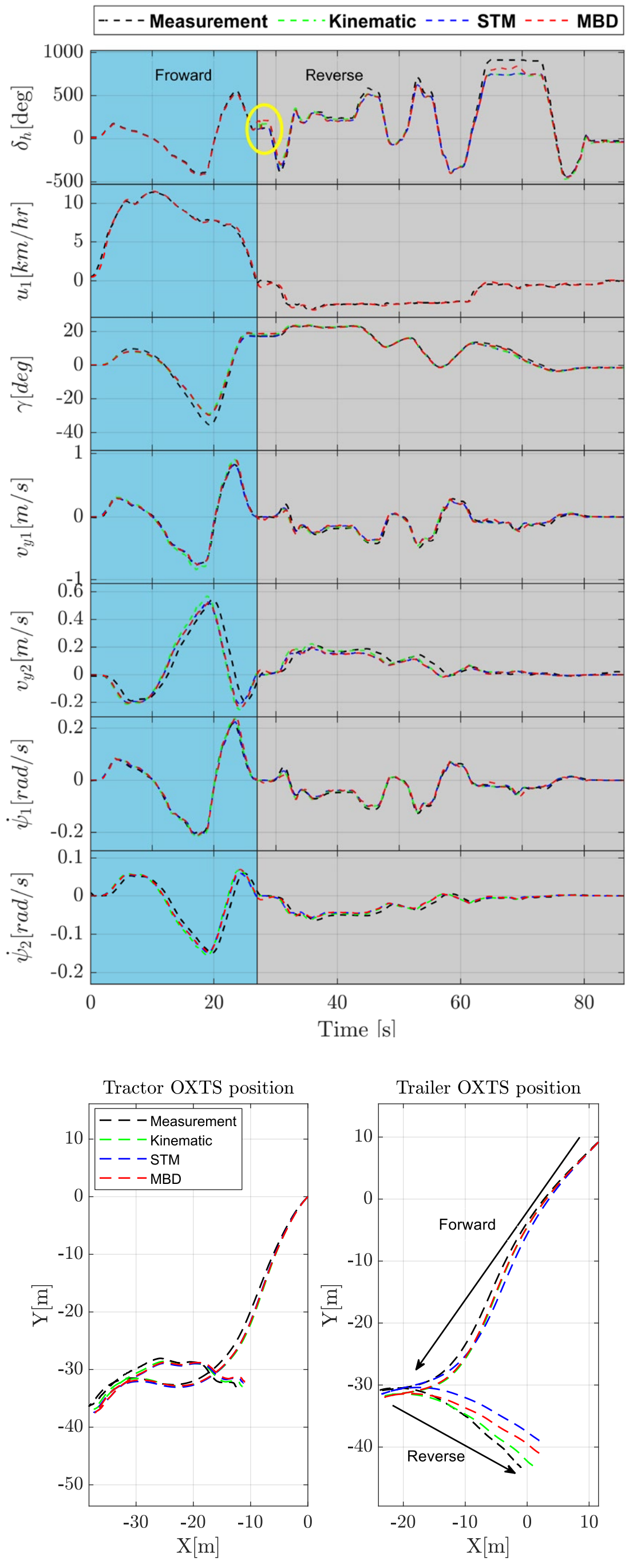

*Fig. 6: Docking Maneuver Test*

### 4.3 Reverse Driving Results

For a qualitative analysis, results of the docking maneuver are represented in Fig. 6. Here, the first two subplots show the steering wheel angle and velocity input to the models. In addition to the yawrate and later velocity subplots, articulation angle has also been shown. It should be noted that the feedback correction in the steering input is only used while reversing, which improves tracking of the articulation angle by the models. Additionally, the implemented control action is not speed dependent hence an abrupt steering effort while shifting from forward to reverse driving (highlighted with yellow circle) can be seen.

Moreover, the accuracy of the model for reverse driving is assessed based on $\bar{\epsilon}_n$ and $J_{steer}$. Because the model deficiency now also appear in the control input. To conclude, a quantitative comparison between the models for the 33 reverse driving maneuvers is listed in Table 3. It can be concluded that the STM model proves to be the most accurate (error 27.7%) with the least steering effort (21.15%). Finally, With the presented approach, we acknowledge that the entire system is being assessed i.e. with feedback control. Therefore, the same controller gain was selected to assess the accuracy of all models.

### 4.4 Computational Efficiency

For efficiency calculation, the KIN and STM model are compared with the MBD model in the simulation environment. The KIN model is 90 % faster and the STM model is 60 % faster than the MBD model. However, the STM model with a reduced number of parameters and state equation is in closer proximity to the measurement results than the KIN model, which makes it an efficient alternative.

## 5. CONCLUSIONS

In this work, we assessed the accuracy of the kinematic, singletrack non-linear and multibody models representing a tractor semi-trailer. The assessment was done by comparing them with real-world measurements that mainly focus on low speed and large steering angle maneuvers. In addition to conventional forward driving model validation, an approach was presented to the reverse driving model validation technique for unstable tractor semi-trailer. A distance based error criterion was defined to evaluate the models. The results from forward and reverse driving conclude that overall the single track non-linear model shows the best performance. In future work, the presented single track vehicle model will be used for path planning and following control.

## ACKNOWLEDGMENT

This work is sponsored by the CATALYST project and the authors would like to thank the following project partners for the support they provided in acquiring the measurement data: DPD Oirschot, TNO automotive and HAN University of applied sciences.

## APPENDIX

| Parameters | Unloaded | Loaded |
|---|---|---|
| Tractor weight, $m_1$ $[kg]$ | 8060 | |
| Trailer weight, $m_2$ $[kg]$ | 7100 | 17360 |
| Tractor inertia, $J_1$ $[kg\ m^2]$ | 11210 | |
| Trailer inertia, $J_2$ $[kg\ m^2]$ | 34613 | 84630 |
| Tractor COG to front axle length, $a_1$ $[m]$ | 1.09 | |
| Tractor COG to drive axle length, $b_1[m]$ | 2.71 | |
| Tractor wheelbase, $L_1[m]$ | 3.8 | |
| Tractor drive axle to fifth wheel length, $L_{1c}$ $[m]$ | 0.67 | |
| Trailer COG to fifth wheel length, $a_2$ $[m]$ | 5.96 | 1.54 |
| Trailer COG to 2nd trailer axle length, $b_2$ $[m]$ | 5.19 | 2.31 |
| Trailer wheelbase $L_2$ $[m]$ | 7.5 | |
| Trailer axle spacing $j$ $[m]$ | 1.3 | |
| Drive axle load, $F_{z2}[N]$ | 34727 | 67100 |
| Trailer axle1 load, $F_{z3}[N]$ | 17658 | 38651 |
| Trailer axle2 load, $F_{z4}[N]$ | 18639 | 39632 |
| Trailer axle3 load, $F_{z5}[N]$ | 18835 | 39436 |
| Normalized cornering stiffness, $f[1/rad]$ | 6 | |
| KIN steering ratio, $i_s$ | 20.5 | |
| STM steering ratio, $i_s$ | 21.5 | |
| MBD steering ratio, $i_s$ | 21 | |
| Steering asymmetry factor, $q$ | 0.1 | |

$$M = \begin{bmatrix} m_1 + m_2 & -m_2(L_c + a_2\cos\gamma) & m_2 a_2 \cos\gamma \\ -m_2(L_c + a_2\cos\gamma) & J_1 + J_2 + m_2(a_2^2 + 2a_2L_c\cos\gamma + L_c^2) & -J_2 - m_2(a_2^2 + a_2L_c\cos\gamma) \\ m_2a_2\cos\gamma & -J_2 - m_2(a_2^2 + a_2L_c\cos\gamma) & J_2 + m_2a_2^2 \end{bmatrix} \quad (11)$$

$$H = \begin{bmatrix} m_1\dot{\psi}_1u_1 + m_2\left(\dot{\psi}_1u_1 - a_2\sin\gamma\dot{\gamma}^2 - a_2\sin\gamma\dot{\psi}_1 + 2a_2\sin\gamma\dot{\gamma}\dot{\psi}_1\right) \\ -m_2(2a_2L_c\sin\gamma\dot{\gamma}\dot{\psi}_1 - a_2L_c\sin\gamma\dot{\gamma}^2 + L_cu_1\dot{\psi}_1 + a_2\cos\gamma\dot{\psi}_1u_1 - a_2\sin\gamma\dot{\psi}_1v_1 \\ m_2a_2\left(\cos\gamma\dot{\psi}_1u_1 - \sin\gamma\dot{\psi}_1v_1 + L_c\sin\gamma\dot{\psi}_1^2\right) \end{bmatrix} \quad (12)$$

$$Q_v = \begin{bmatrix} F_{y1}\cos\delta + F_{y2} + \left(F_{y3} + F_{y4} + F_{y5}\right)\cos\gamma \\ a_1F_{y1}\cos\delta - b_1F_{y2} - L_c\left(F_{y3} + F_{y4} + F_{y5}\right)\cos\gamma - L_{21}F_{y3} - L_{22}F_{y4} - L_{23}F_{y5} \\ L_{21}F_{y3} + L_{22}F_{y4} + L_{23}F_{y5} \end{bmatrix} \quad (13)$$

where,

$$F_{yi} = fF_{zi}\alpha_i \quad (14)$$